\documentclass[journal]{IEEEtran}
\usepackage{amsmath}
\usepackage{graphicx}
\usepackage{url}
\usepackage{hyperref}
\usepackage{cite} 

\begin{document}

\title{Evaluating SAT and SMT Solvers on Large-Scale Sudoku Puzzles}

\author{
    \IEEEauthorblockN{Liam Davis\textsuperscript{1}, Tairan ``Ryan'' Ji\textsuperscript{1}}\\
    \IEEEauthorblockA{\textsuperscript{1}Amherst College, Amherst, MA, USA\\
    Emails: \{ljdavis27, tji26\}@amherst.edu}
}

\maketitle

\begin{abstract}
Modern SMT solvers have revolutionized the approach to constraint satisfaction problems by integrating advanced theory reasoning and encoding techniques. In this work, we evaluate the performance of modern SMT solvers in Z3, CVC5 and DPLL(T) against a standard SAT solver in DPLL. By benchmarking these solvers on novel, diverse 25x25 Sudoku puzzles of various difficulty levels created by our improved Sudoku generator, we examine the impact of advanced theory reasoning and encoding techniques. Our findings demonstrate that modern SMT solvers significantly outperform classical SAT solvers. This work highlights the evolution of logical solvers and exemplifies the utility of SMT solvers in addressing large-scale constraint satisfaction problems.
\end{abstract}


\section{Introduction}
Sudoku is a well-known puzzle that has been around for many decades. Due to its structure, it can be understood as a type of constraint satisfaction problem, suitable to be solved by SAT and SMT solvers. SAT solvers, like the DPLL algorithm, are effective at solving problems by encoding them with boolean logic. In contrast, SMT solvers extend this capability, enabling them to handle a broader range of constraints. Modern SMT solvers allow for more creative and efficient encodings that lead to more efficient solvers. In this paper, we test modern SMT solvers (Z3, CVC5, and DPLL(T)) against the DPLL algorithm on a benchmark of novel, diverse 25x25 Sudoku puzzles created by our generator to answer a few different questions. First, how have logical solvers evolved over time in terms of performance and capability? Second, how do different encodings of Sudoku affect the efficiency and scalability of these solvers? Lastly, are there specific features or optimizations in SMT solvers that provide a significant advantage over traditional SAT solvers for this class of problem?

\section{Motivation and Background}
Sudoku puzzles, particularly larger variants such as 25x25 grids, present a significant computational challenge. This makes solving such puzzles an excellent case study for evaluating the performance of logical solvers. Traditional SAT solvers, like the DPLL algorithm, are efficient at handling problems defined by Boolean logic. But, they are less efficient when a problem has arithmetic and equality constraints. 

SMT solvers overcome these limitations by integrating theory solvers, which extend SAT solving capabilities to richer domains such as linear arithmetic, arrays, and equality with uninterpreted functions. Thus, SMT allows for more expressive encodings, and often more efficient problem-solving strategies.

The motivation for this work arises from the growing importance of constraint satisfaction problems in real-world applications, such as scheduling, planning, and verification. By exploring the evolution of logical solvers, this project aims to identify how modern SMT solvers have improved over traditional SAT solvers in terms of performance, scalability, and capability on large-scale constraint satisfaction problems.

\section{Tool and Code Availability}
As part of this project, we developed a Python-based tool for benchmarking SMT and SAT solvers by evaluating their performance on 25x25 Sudoku puzzles. The tool provides implementations of the solvers evaluated in this paper, as well as a generator that creates Sudoku puzzles of various difficulty levels, a Depth First Search (DFS)-based solver used during the generation process, and a framework for benchmarking the SMT and SAT solvers on the generated puzzles.

The code, including benchmarks and experiments, is publicly available on GitHub at 
\href{https://github.com/liamjdavis/Sudoku-SMT-Solvers}{github.com/liamjdavis/Sudoku-SMT-Solvers}. 
The corresponding Python package is published on PyPI at 
\href{https://pypi.org/project/sudoku-smt-solvers/}{pypi.org/project/sudoku-smt-solvers/}. 
The documentation is available at 
\href{https://liamjdavis.github.io/Sudoku-SMT-Solvers/}{liamjdavis.github.io/Sudoku-SMT-Solvers}. 
The project is open source and distributed under the MIT license. We welcome contributions from the community to enhance the tool, add new features, and extend its capabilities.

\section{Methodology}

\subsection{Baseline Implementation}

The baseline implementation provides a framework to encode and solve Sudoku puzzles, built around the following key components:

\begin{itemize}
    \item \textbf{Input Validation:} Each solver verifies that the input is a 25x25 grid of integers, with each cell containing a value between 0 and 25 (where 0 represents an empty cell). Invalid inputs raise errors to prevent undefined behavior.
    
    \item \textbf{Propagation Statistics:} Solvers maintain propagation statistics that provide further insight on performance.
    
    \item \textbf{Encoding Rules of Sudoku:}
    \begin{itemize}
        \item \textbf{Cell Constraints:} Each cell in the grid must have exactly one value between 1 and 25. 
        \item \textbf{Row, Column, and Block Constraints:} Logical constraints ensure that no number appears more than once in any row, column, or 5x5 block.
        \item \textbf{Symmetry-Breaking Constraints:} Additional constraints are added to reduce redundant search by eliminating equivalent solutions.
    \end{itemize}

    \item \textbf{Solution Extraction and Validation:} 
    Once the solving process is complete, variable assignments are mapped back to the Sudoku grid. The solution is validated to ensure correctness and alignment with the initial fixed values.

\end{itemize}

\subsection{DPLL Solver}

The DPLL solver processes the Sudoku puzzle by encoding it into Conjunctive Normal Form (CNF) and systematically solving it using the PySAT Python package. The solver integrates CNF generation, efficient SAT solving techniques, and solution validation to handle large puzzles like 25x25 Sudoku.

\subsubsection{Overview of the DPLL Algorithm}

The DPLL algorithm employs a backtracking-based approach for solving Boolean satisfiability (SAT) problems. It operates on CNF formulas and systematically explores the space of variable assignments to generate a satisfying solution. The key steps of the algorithm include:

\begin{enumerate} 
    \item \textbf{Unit Propagation:} Simplifies the formula by assigning values to variables that must hold true for a clause to be satisfied. This process iteratively reduces the formula’s complexity.
    \item \textbf{Pure Literal Elimination:} Identifies literals that appear with only one polarity (either positive or negative) in the formula and assigns them values that satisfy all clauses containing them.
    \item \textbf{Backtracking Search:} Recursively assigns truth values to variables. When a conflict is detected (i.e., a clause cannot be satisfied), the algorithm backtracks to try alternative assignments.
\end{enumerate}

The DPLL algorithm forms the foundation for modern SAT solvers, incorporating these fundamental techniques alongside heuristics to improve performance.

\subsubsection{Implementation Details}

The specific implementation of the DPLL solver for Sudoku includes the following components:

\paragraph{CNF Generation} 
The Sudoku puzzle is converted into a CNF formula, where:
\begin{itemize}
    \item Each cell is represented as a set of Boolean variables, one for each potential value (1–25).
    \item Constraints ensure that each cell contains exactly one value, and each row, column, and 5x5 block contains distinct values.
    \item Pre-filled cells are encoded as fixed assignments to enforce consistency with the initial puzzle.
\end{itemize}

\paragraph{SAT Solver Integration} 
The solver uses the PySAT library's low-level SAT solving capabilities to handle Boolean clause propagation, recursive search, and conflict resolution. Key techniques include:
\begin{itemize}
    \item \textbf{Unit Clause Propagation:} Automatically handled by the SAT solver to simplify clauses during solving.
    \item \textbf{Backtracking:} The solver explores alternative variable assignments when conflicts arise, ensuring exhaustive search within the solution space.
\end{itemize}

\paragraph{Solution Extraction and Validation} 
If the SAT solver finds a satisfying assignment, the model is decoded into a 2D Sudoku grid. The solution is validated to ensure it satisfies all Sudoku constraints, including unique values in each row, column, and block.

\subsubsection{Summary}

The DPLL solver serves as a baseline for evaluating more advanced solvers, leveraging efficient SAT-solving techniques to solve large constraint satisfaction problems.

\subsection{DPLL(T) Solver}

The DPLL(T) solver builds upon the traditional DPLL algorithm by incorporating theory-specific reasoning, enabling it to handle richer constraints such as arithmetic and equality. This approach is based on the framework presented by Nieuwenhuis, Oliveras, and Tinelli in their foundational work on DPLL(T) \cite{Nieuwenhuis2006}. The implementation uses the PySAT Python package and integrates low-level SAT solving capabilities with Sudoku-specific theory reasoning.

\subsubsection{Overview of the DPLL(T) Algorithm}

The DPLL(T) algorithm extends the core principles of DPLL with theory-specific reasoning to enhance its solving capabilities. Key components of the algorithm include:

\begin{enumerate} 
    \item \textbf{Theory Propagation:} In addition to unit propagation for Boolean clauses, the solver propagates constraints based on domain-specific theories (e.g., arithmetic, equality, or Sudoku-specific rules). This integration ensures invalid assignments are detected early.
    \item \textbf{Theory Conflict Resolution:} When a conflict is identified within the theory, the solver generates theory-specific conflict clauses that are added to the Boolean formula. These clauses prevent revisiting invalid states and streamline the search process.
    \item \textbf{Incremental Solving:} DPLL(T) interleaves SAT-solving steps with theory-specific checks, iteratively refining the solution space by combining Boolean and theory-level deductions.
    \item \textbf{Clause Learning and Backtracking:} The algorithm incorporates learned clauses to reduce redundant exploration and employs intelligent backtracking to efficiently explore alternative assignments.
\end{enumerate}

\subsubsection{Implementation Details}

The DPLL(T) solver for Sudoku incorporates several specialized components:

\paragraph{Theory Propagation} 
The solver dynamically enforces Sudoku-specific constraints during the solving process:
\begin{itemize}
    \item Checks for violations of row, column, and block uniqueness.
    \item Identifies conflicts arising from invalid partial assignments and generates conflict clauses to prevent revisiting the same invalid states.
\end{itemize}

\paragraph{Conflict Analysis and Clause Learning} 
When a conflict is detected during theory propagation, the solver generates a conflict clause that captures the root cause of the conflict. This clause is added to the formula to guide subsequent search steps and avoid redundant conflicts.

\paragraph{SAT Clause Management} 
The solver leverages the PySAT library's SAT solving capabilities to handle Boolean clause propagation, conflict detection, and clause learning:
\begin{itemize}
    \item Original clauses represent the problem's constraints (e.g., cell, row, column, and block rules).
    \item Learned clauses, generated during theory propagation, improve the solver's efficiency by reducing redundant search.
\end{itemize}

\paragraph{CNF Encoding} 
The Sudoku puzzle is encoded as a CNF (Conjunctive Normal Form) formula:
\begin{itemize}
    \item Each cell is represented as a Boolean variable encoding its potential values.
    \item Constraints enforce that each cell contains exactly one value, and all rows, columns, and 5x5 blocks contain distinct numbers.
    \item Pre-filled cells are encoded as fixed assignments to ensure consistency with the initial puzzle.
\end{itemize}

\paragraph{Model Extraction and Validation} 
If the solver finds a satisfying assignment, the Boolean model is decoded to extract a Sudoku solution. The solution is validated to ensure it satisfies all constraints.

\subsubsection{Summary}

The DPLL(T) solver demonstrates the advantages of combining low-level SAT-solving techniques with high-level domain-specific reasoning. This hybrid approach extends the DPLL solver to achieve greater scalability for solving large constraint satisfaction problems, such as 25x25 Sudoku puzzles.

\subsection{Z3 Solver}

The Z3 solver leverages the SMT (Satisfiability Modulo Theories) capabilities of the Z3 library to solve 25x25 Sudoku puzzles. It incorporates advanced techniques such as integer variable encoding, efficient constraint propagation, and model evaluation. This solver uses the Z3 Python package, originally developed by Leonardo Mendon{\c{c}}a de Moura and Nikolaj Bj{\o}rner \cite{deMoura2008}.

\subsubsection{Overview of Z3}

SMT solving generalizes SAT solving by incorporating theories that model specific domains. The Z3 solver extends foundational SAT-solving techniques with theory solvers, enabling it to handle richer constraints. Key components include:

\begin{enumerate}
    \item \textbf{Theory Integration:} Theories are tightly integrated into the solving process, enabling the solver to handle constraints involving domain-specific properties (e.g., integer arithmetic for Sudoku).
    \item \textbf{Efficient Propagation and Conflict Resolution:} The solver combines Boolean reasoning with theory propagation to quickly identify conflicts and resolve them using learned constraints.
    \item \textbf{Model Evaluation:} If a satisfiable assignment is found, Z3 extracts the model, representing the solution in terms of the original variables and their domains.
    \item \textbf{Incremental Solving:} Z3 allows constraints to be added or removed dynamically, making it versatile for solving problems that evolve over time.
\end{enumerate}

\subsubsection{Implementation Details}

The Z3 solver applies SMT-solving principles to encode and solve Sudoku puzzles efficiently. The implementation includes the following components:

\paragraph{Variable Encoding}
Each cell in the Sudoku grid is represented by a Z3 integer variable with a domain of 1–25. A 2D array of Z3 integer variables is created, where:
\begin{itemize}
    \item The variable \texttt{x\_i\_j} represents the value in the cell at row \(i\), column \(j\).
\end{itemize}

\paragraph{Encoding Sudoku Rules}
The solver encodes Sudoku rules as logical constraints using Z3's API:
\begin{itemize}
    \item \textbf{Cell Constraints:} Each cell must contain a value between 1 and 25:
    \[
    1 \leq x\_i\_j \leq 25
    \]
    \item \textbf{Row Constraints:} All values in each row must be distinct, enforced using Z3's \texttt{Distinct} function:
    \[
    \texttt{Distinct}(\{x\_i\_1, x\_i\_2, \ldots, x\_i\_{25}\})
    \]
    \item \textbf{Column Constraints:} All values in each column must be distinct:
    \[
    \texttt{Distinct}(\{x\_1\_j, x\_2\_j, \ldots, x\_{25}\_j\})
    \]
    \item \textbf{Block Constraints:} All values in each 5x5 subgrid must be distinct. Subgrid constraints are systematically applied for all subgrids.
\end{itemize}

\paragraph{Encoding the Puzzle}
The initial Sudoku puzzle is encoded by asserting equality constraints for cells with predefined values. For example, if the value \(k\) is already present in cell \((i, j)\), the solver asserts:
\[
x\_i\_j = k
\]

\paragraph{Theory Propagation and Model Checking}
Once constraints are added, the Z3 solver performs theory propagation and searches for a satisfying assignment using SMT techniques:
\begin{itemize}
    \item The \texttt{check()} method determines whether the puzzle is satisfiable.
    \item If satisfiable, the \texttt{model()} method extracts the solution.
\end{itemize}

\subsubsection{Summary}

The Z3 solver highlights the power of SMT techniques in handling complex constraint satisfaction problems. By combining integer variable encoding, efficient constraint propagation, and advanced model evaluation, it achieves scalability and performance for solving large Sudoku puzzles, including 25x25 grids.

\subsection{CVC5 Solver}

The CVC5 solver leverages its SMT capabilities to solve 25x25 Sudoku puzzles using integer variable encoding, logical constraint propagation, and model evaluation. It is implemented using the CVC5 Python package, originally developed by Clark Barrett et al. \cite{DBLP:conf/tacas/BarbosaBBKLMMMN22}.

\subsubsection{Overview of CVC5}

CVC5 builds on SAT-solving foundations, extending them with domain-specific reasoning via theories. The solver interleaves Boolean reasoning with theory-specific propagation and conflict resolution to explore the solution space. Key features include:

\begin{enumerate}
    \item \textbf{Theory Integration:} Theories such as linear integer arithmetic are integrated into the solving process, allowing direct handling of domain-specific constraints.
    \item \textbf{Efficient Conflict Resolution:} CVC5 resolves conflicts dynamically by combining Boolean and theory-level reasoning to guide the search process.
    \item \textbf{Incremental Solving and Model Evaluation:} Constraints can be added incrementally, and satisfying assignments (models) are evaluated to extract solutions.
    \item \textbf{Quantifier-Free Linear Integer Arithmetic (QF\_LIA):} 
    CVC5 leverages this theory to efficiently represent and solve the integer constraints required for Sudoku.
\end{enumerate}

\subsubsection{Implementation Details}

The CVC5 solver applies SMT-solving principles to encode and solve Sudoku puzzles. The implementation includes the following components:

\paragraph{Variable Creation}
Each cell in the Sudoku grid is represented by a CVC5 integer variable with a domain of 1–25. A 2D array of variables is created, where:
\begin{itemize}
    \item The variable \texttt{x\_i\_j} represents the value in the cell at row \(i\), column \(j\).
    \item The solver is configured for Quantifier-Free Linear Integer Arithmetic and supports model production and incremental solving.
\end{itemize}

\paragraph{Encoding Sudoku Rules}
The solver encodes the rules of Sudoku as logical constraints:
\begin{itemize}
    \item \textbf{Domain Constraints:} Each variable must be between 1 and 25.
    \item \textbf{Row Constraints:} Each row contains distinct values, enforced using CVC5's \texttt{DISTINCT} operator.
    \item \textbf{Column Constraints:} Each column contains distinct values.
    \item \textbf{Block Constraints:} Each 5x5 subgrid contains distinct values, systematically encoded by grouping variables within the subgrid.
\end{itemize}

\paragraph{Encoding the Puzzle}
The initial Sudoku puzzle is encoded by asserting equality constraints for cells with predefined values. For example, if the value \(k\) is already present in cell \((i, j)\), the solver asserts:
\[
\texttt{x\_i\_j = k}
\]
using the \texttt{assertFormula} method.

\paragraph{Theory Propagation and Model Checking}
After encoding the constraints, the solver uses SMT-based theory propagation and conflict detection:
\begin{itemize}
    \item The \texttt{checkSat()} method determines whether the puzzle is satisfiable.
    \item If satisfiable, the \texttt{getValue()} method retrieves the solution by evaluating the model.
\end{itemize}

\paragraph{Solution Extraction}
The solution is extracted by evaluating the integer variables for each cell. The values are assigned to the Sudoku grid, and the solution is validated to ensure it satisfies all constraints.

\subsubsection{Summary}

The CVC5 solver demonstrates the power of modern SMT techniques in solving large constraint satisfaction problems. By integrating integer variable encoding, efficient constraint propagation, and advanced model evaluation, it provides robust performance and flexibility for solving complex puzzles such as 25x25 Sudoku grids.

\subsection{Benchmark Suite}

In addition to the solvers, we also implemented a benchmark suite designed to run the solvers on various 25x25 Sudoku puzzles efficiently and systematically. The benchmark suite includes the following key features:

\begin{itemize}
    \item \textbf{Configurable Timeout Handling:} The benchmark suite enforces a user-defined timeout for each puzzle, ensuring that execution is terminated if the solver exceeds the specified time limit. 
    
    \item \textbf{Detailed Metric Collection:} The benchmark suite tracks performance metrics for each solver run. These metrics include the solution status, the total solving time, and the number of clause propagations performed.

    \item \textbf{Automated Result Aggregation:} Performance data is automatically consolidated across solvers. The suite computes aggregate statistics such as the total number of puzzles solved, the average solving time, and the average clause propagation counts.

    \item \textbf{Result Exporting:} Benchmark results are saved in timestamped CSV files. 
\end{itemize}

The benchmark suite serves as a versatile and reliable tool for evaluating solver performance under controlled conditions. It enables rigorous comparisons of different solvers.

\subsection{Development and Testing Practices}

To ensure the reliability, maintainability, and scalability of our solvers, we adopted modern software engineering practices throughout the development process. These practices included:

\begin{itemize}
    \item \textbf{Automated CI/CD Pipelines:} 
    We used GitHub Actions to automate the linting, testing, and code coverage verification processes. Two workflows were implemented:
    \begin{itemize}
        \item \textit{Styling Workflow:} This workflow runs on every push and pull request, ensuring that the codebase adheres to Python's PEP 8 style guidelines.
        \item \textit{Pytest Workflow:} This workflow executes unit and integration tests with \texttt{pytest} and measures code coverage using \texttt{pytest-cov}. This workflow is ran for python versions 3.10, 3.11 and 3.12.
        \item \textit{CodeQL Workflow:} 
        To enhance code security and quality, we integrated a CodeQL analysis workflow that identifies potential vulnerabilities in the code.
    \end{itemize}

    \item \textbf{Code Formatting and Pre-commit Hooks:}
    We enforced consistent code style using the Black formatter with pre-commit hooks.

    \item \textbf{Test Coverage Standards:}
    Code coverage is measured using pytest with a required coverage of 90\% code coverage, as reported by Coveralls.

    \item \textbf{Test Suite Structure:}
    The test suite was designed to be automatically managed by our pytest configuration.
\end{itemize}

By incorporating these practices, we maintained a high-quality codebase, streamlined development workflows, and ensured reproducibility in testing and benchmarking. In the end, we achieved these key statistics:

\begin{itemize}
    \item \textbf{94\% Code Coverage}
    \item \textbf{73 Unit Tests}
\end{itemize}

\section{Evaluation Methodology}

\subsection{Difficulty Level Classification}
To evaluate the performance of different solvers, we generated a dataset of 100 25x25 Sudoku puzzles categorized by five difficulty levels, from extremely easy to evil. Two metrics were used to determine the difficulty level of the puzzles:
\begin{enumerate}
    \item \textbf{Number of Cells Given:}  
    A greater number of cells already filled in (``given'') in the initial puzzle reduces the number of choices the solver has to make, so puzzles with fewer givens should be rated as more difficult.

    \item \textbf{Lower Bound on the Number of Given Cells in Each Row and Column:}
    The distribution of given cells across a puzzle's grid can significantly affect the puzzle's difficulty. Specifically, given two puzzles with the same number of total givens, the one with a higher lower bound is easier to solve because the bound forces a more uniform distribution of givens across the grid, which increases the total number of constraints the solvers can utilize and reduces the search space.  
\end{enumerate}

The metrics described above are selected from a set of Sudoku evaluation metrics developed by Yuan-Hai et al. \cite{2009Sudoku}. Given the specifications of our project, we selected the two metrics most relevant for determining solving difficulty for SMT and SAT solvers and modified their difficulty level mappings so they can be applied to 25x25 grids. The modified mapping is described in the table below:

\begin{table}[h]
\centering
\caption{Difficulty Level Mapping}
\label{tab:diff_mapping}
\begin{tabular}{|l|c|c|}
\hline
\textbf{Level}          & \textbf{Number Givens} & \textbf{Lower Bound (Row/Col)} \\ \hline
Extremely Easy          & $382$--$\infty$           & 14                             \\ \hline
Easy                    & $274$--$381$              & 11                             \\ \hline
Medium                  & $243$--$273$              & 7                              \\ \hline
Difficult               & $212$--$242$              & 4                              \\ \hline
Evil                    & $166$--$211$              & 0                              \\ \hline
\end{tabular}
\end{table}

Following the difficulty mapping defined in the table above, we significantly improved upon the algorithm described by Yuan-Hai et al. \cite{2009Sudoku} to generate the 100 puzzles used by our benchmark.

\subsection{Sudoku Generator}

In their paper, Yuan-Hai et al. \cite{2009Sudoku} described an algorithm they developed for generating diverse 9x9 Sudoku puzzles of desired difficulties by first creating a "terminal pattern" (a completely filled-in grid that adheres to the Sudoku game rules) and then digging holes (removing the value in certain cells). We first implemented this algorithm based on descriptions of it in the paper, then generalized it to allow for the generation of puzzles of any size and made substantial optimizations to improve the runtime. Our final algorithm is as follows:

\begin{enumerate}
    \item \textbf{Terminal Pattern Generation:}
    As Yuan-Hai et al. \cite{2009Sudoku} describes, terminal patterns should be generated using the Las Vegas Algorithm to ensure the diversity of the generated puzzles:
    \begin{enumerate}
        \item Starting with an empty grid, choose $n$ random cells to be used as ``seeds"
        \item Fill in the $n$ seed cells with some combination of values such that all Sudoku rules are satisfied; if no such combinations exist, return to step 1.a
        \item Run a Depth First Search (DFS) Solver on the seeded grid until either a solution/terminal pattern is found (proceed to step 2) or a timeout is hit (return to step 1.a)
    \end{enumerate}
    Our implementation of terminal pattern generation and DFS solver utilizes algorithmic techniques not specified in the Yuan-Hai et al. paper, such as the Minimum Remaining Values (MRV) heuristic, to improve efficiency. In addition, we have found that setting the number of seeds $n$ to 80 and the timeout to 5 seconds significantly reduces the time it takes to generate a terminal pattern.
    \item \textbf{Hole Digging:}
    Starting with a terminal pattern, a Sudoku puzzle of the desired difficulty level can be generated by digging holes in the grid (i.e., erasing the values in certain cells) \cite{2009Sudoku}:
    \begin{enumerate}
        \item Choose difficulty level
        \item Determine the ``digging pattern," which corresponds to the order in which cells are examined as potential holes. Based on the analysis made by Yuan-Hai et al., to maximize the diversity of produced puzzles while minimizing runtime, the sequence should: 
        \begin{itemize}
            \item Be randomized globally for \textbf{extremely easy} and \textbf{easy} puzzles
            \item Start at the top left cell and cover every other cell for \textbf{medium} puzzles
            \item Start at the top left cell and follow an "S" pattern for \textbf{difficult} puzzles
            \item Start at the top left cell and proceed from left to right, then top to bottom for \textbf{evil} puzzles
        \end{itemize}
        For more details on the patterns, refer to the Yuan-Hai et al. paper \cite{2009Sudoku}.
        \item Based on the selected difficulty level, set the minimum number of givens the puzzle must have and the lower bound on the number of givens in each row and column. These two numbers are uniformly selected from the ranges specified in Table~\ref{tab:diff_mapping}.
        \item Set all cells in the grid to "can-be-dug".
        \item Explore the first cell in the digging sequence by checking if it can be dug (digging will not violate the numbers set in step 2.c) and if digging it will result in a puzzle with a unique solution. In the Yuan-Hai et al. paper, uniqueness was checked for using a novel reduction-to-absurdity-based method that checked using the DFS solver whether the puzzle had a solution if the value in the cell was replaced with each of the 24 other possible values; if no other value resulted in a puzzle with a solution, digging the hole will create a unique puzzle. However, we discovered that it is significantly more efficient to remove the value in the cell and run the DFS solver on the resulting puzzle; as soon as the solver finds two solutions, we can terminate early with the knowledge that digging the hole will not create a unique puzzle. We estimate that our algorithm to check for uniqueness is roughly \textbf{24 times faster per cell checked} compared to Yuan-Hai et al.'s algorithm.
        \item If the first cell can be dug and if digging will result in a unique puzzle, dig the cell. Then, regardless of whether the cell is dug or not, perform pruning by removing the cell from the list of "can-be-dug" cells and explore the next cell in the sequence. Keep exploring until no "can-de-dug" cells exist.
    \end{enumerate}
\end{enumerate}

\subsection{Experimental Process}

To evaluate the solvers' performance systematically, we followed these steps:

\begin{enumerate}
    \item \textbf{Benchmark Selection:}  
    We generated a benchmark set of 100 Sudoku puzzles with the method described earlier that included:  
    \begin{itemize}
        \item 20 \textbf{Extremely Easy} puzzles  
        \item 20 \textbf{Easy} puzzles  
        \item 20 \textbf{Medium} puzzles  
        \item 20 \textbf{Difficult} puzzles  
        \item 20 \textbf{Evil} puzzles  
    \end{itemize}

    \item \textbf{Solver Configuration:}  
    Each solver was tested using its default configuration with no additional tuning or optimization.

    \item \textbf{Execution Environment:}  
    All experiments were conducted in the same execution environment.

    \item \textbf{Testing Procedure:}  
    \begin{itemize}
        \item Each solver was ran on the set of 100 benchmarks sequentially.
        \item A timeout of 30 seconds was imposed on each puzzle.
    \end{itemize}

\end{enumerate}

\subsection{Performance Metrics}
To compare the solvers, we collected the following performance metrics:
\begin{itemize}
    \item \textbf{Success Rate:} The percentage of puzzles solved correctly within the timeout.
    \item \textbf{Solving Time:} The total time taken by a solver to find a solution within the timeout.
    \item \textbf{Propagation Efficiency:} The number of clauses or constraints propagated per second, as reported by the solver's internal statistics.
\end{itemize}
    
These metrics were analyzed to identify trends in solver performance across difficulty levels. The evaluation process provided insights into how different solvers match up on various difficulty categories, highlighting their strengths and weaknesses. This comprehensive analysis helps determine the most effective solver for each class of Sudoku puzzles.

\section{Results}

This section presents the performance evaluation of the solvers across key metrics: success rates, solving times, and propagation efficiency. The results are analyzed to provide a comprehensive understanding of each solver's capabilities.

\subsection{Solver Success Rates}

Table \ref{tab:success_rates} summarizes the success rates of each solver, showing the percentage of puzzles solved correctly within the timeout.

\begin{table}[h]
\caption{Solver Success Rates}
\label{tab:success_rates}
\centering
\begin{tabular}{|l|c|c|c|}
\hline
\textbf{Solver} & \textbf{Total Puzzles} & \textbf{Solved Count} & \textbf{Success Rate (\%)} \\ \hline
CVC5            & 100                    & 100                   & 100.0                      \\ 
DPLL            & 100                    & 97                    & 97.0                       \\ 
DPLL(T)         & 100                    & 97                    & 97.0                       \\ 
Z3              & 100                    & 100                   & 100.0                      \\
\hline
\end{tabular}
\end{table}

The success rates indicate that CVC5 and Z3 solved every puzzle within the specified timeout, demonstrating their robustness in handling large and complex Sudoku puzzles. In contrast, DPLL and DPLL(T) encountered failures in 3\% of the cases. This highlights the limitations of classical approaches when applied to large-scale problems, compared to modern SMT solvers like Z3 and CVC5.

\subsection{Average Solve Time}

\begin{figure}[h!]
\centering
\includegraphics[width=0.5\textwidth]{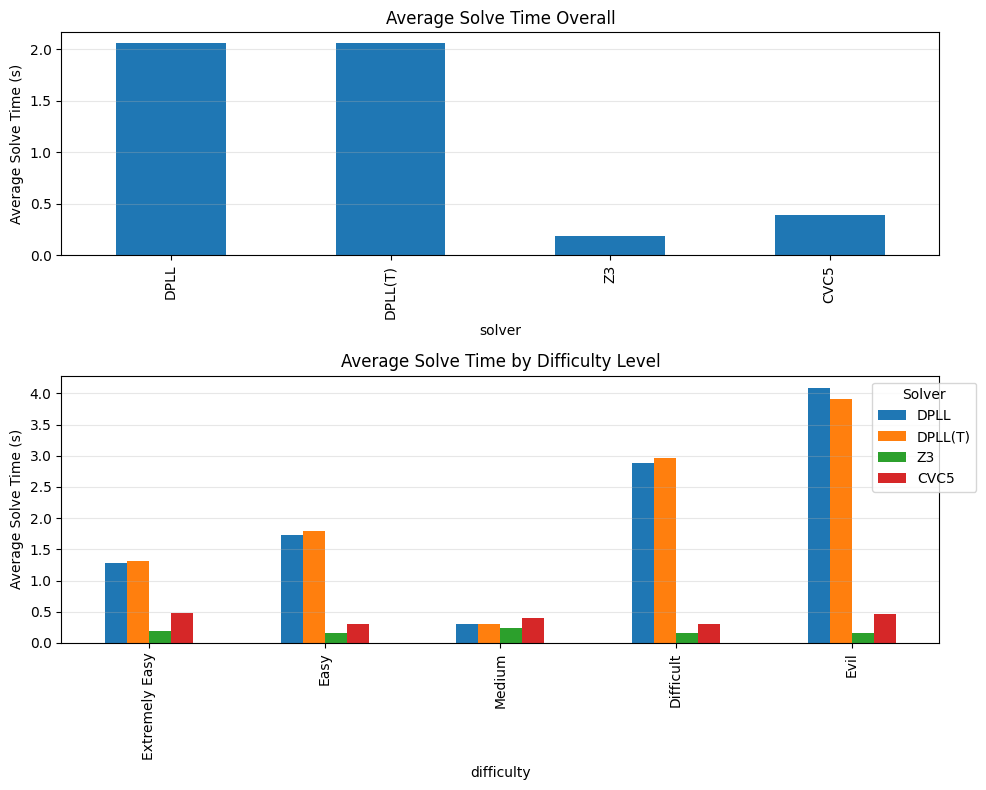}
\caption{Average Solving Time for Each Solver}
\label{fig:avg_solve_time}
\end{figure}

Figure~\ref{fig:avg_solve_time} illustrates the average solving time for each solver across all puzzles.

The results show that Z3 achieved the lowest average solving time, closely followed by CVC5. Both solvers demonstrate the efficiency of modern SMT techniques, which combine Boolean reasoning with domain-specific knowledge to streamline problem-solving. DPLL and DPLL(T), on the other hand, exhibited significantly higher solving times. The stark difference in performance underscores the benefits of integrating domain-specific reasoning in SMT solvers.

\subsection{Solve Time Scatter Plot}

\begin{figure}[h!]
\centering
\includegraphics[width=0.4\textwidth]{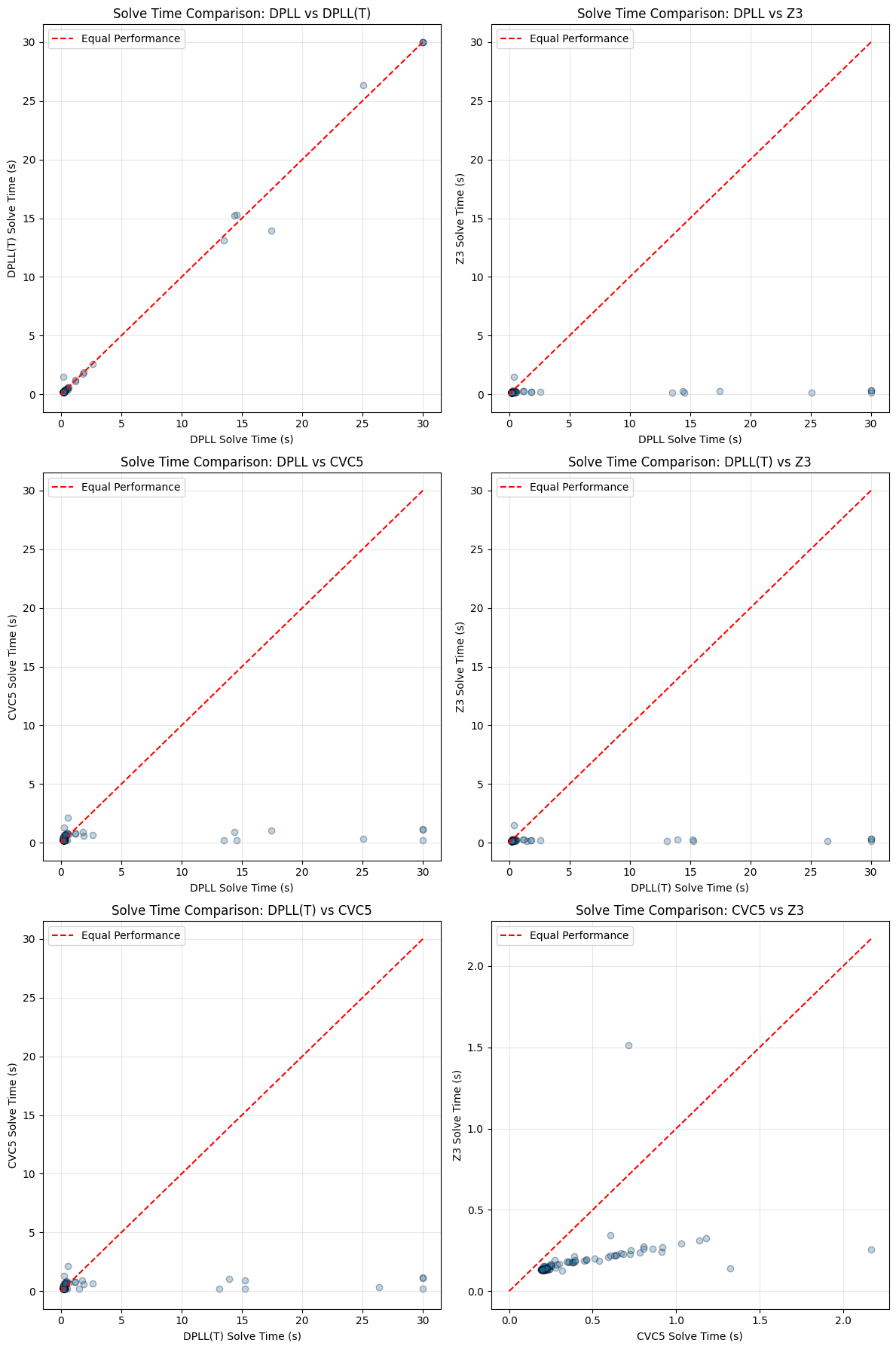}
\caption{Scatter Plots of Solve Times Across Solvers}
\label{fig:scatter_solve_time}
\end{figure}

Figure~\ref{fig:scatter_solve_time} presents scatter plots of solving times for each puzzle, providing a direct comparison of the solvers' performance.

The scatter plots reveal that DPLL and DPLL(T) exhibit nearly identical performance patterns, suggesting that the addition of theory-specific reasoning in DPLL(T) provided marginal gains in this context. Meanwhile, CVC5 and Z3 consistently outperform the classical solvers, with Z3 maintaining a slight edge over CVC5 in most cases. This comparison further emphasizes the superiority of SMT solvers in tackling constraint satisfaction problems efficiently, particularly in scenarios where rapid conflict resolution and optimized propagation are critical.

\subsection{Average Propagations}

\begin{figure}[h!]
\centering
\includegraphics[width=0.4\textwidth]{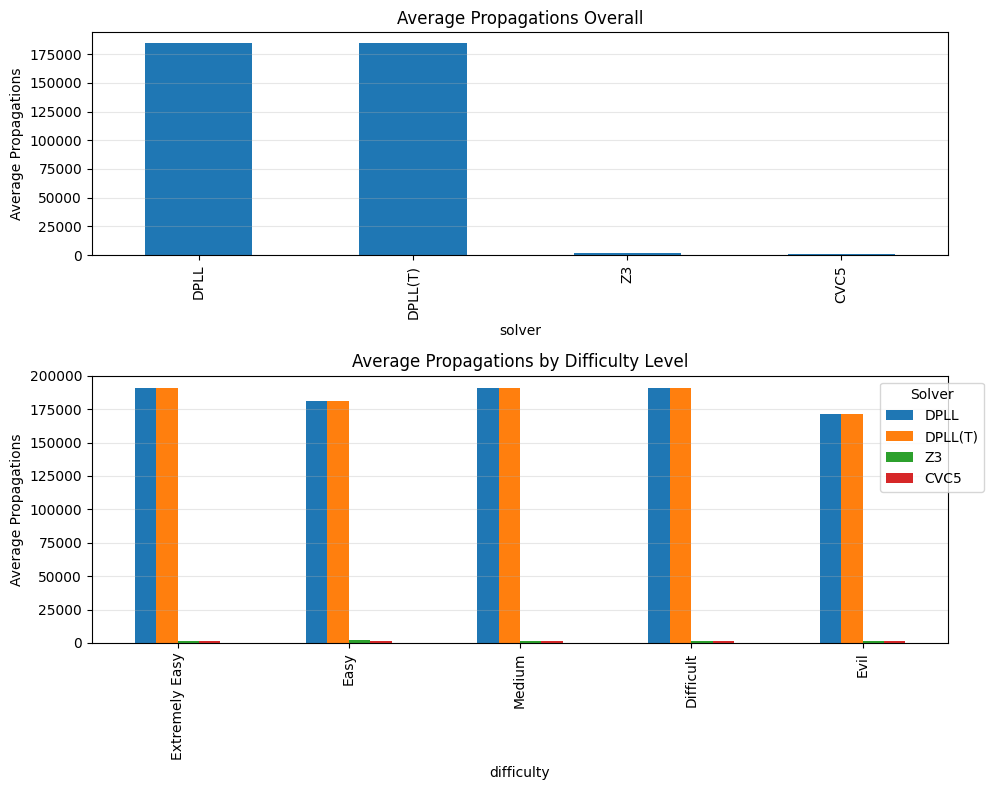}
\caption{Average Propagations}
\label{fig:avg_propagations}
\end{figure}

Figure~\ref{fig:avg_propagations} highlights the propagation efficiency of the solvers, showing the average number of propagations required to solve a single puzzle.

CVC5 and Z3 demonstrate significantly fewer propagations per puzzle compared to DPLL and DPLL(T). This indicates that modern SMT solvers leverage more efficient propagation mechanisms and conflict resolution strategies, allowing them to reach solutions with fewer computational steps. Notably, CVC5 achieved the best propagation efficiency, showcasing its strength in managing constraints and reducing redundant computations. In contrast, the high propagation counts for DPLL and DPLL(T) reflect their reliance on less optimized techniques, resulting in longer solving times and reduced efficiency.

\subsection{Conclusion of Results}

The results clearly indicate that modern SMT solvers, particularly CVC5 and Z3, outperform a classical SAT solver in DPLL and a legacy SMT solver in DPLL(T) in solving large 25x25 Sudoku puzzles. This performance gap is evident across all evaluated metrics:

\begin{itemize}
    \item \textbf{Success Rates:} CVC5 and Z3 solved 100\% of the puzzles, while DPLL and DPLL(T) encountered failures in 3\% of cases.
    \item \textbf{Solving Times:} Z3 and CVC5 demonstrated significantly lower solving times, leveraging advanced SMT techniques for efficient problem-solving.
    \item \textbf{Propagation Efficiency:} CVC5 exhibited the fewest propagations, followed closely by Z3, highlighting their ability to manage constraints and reduce redundant computations effectively.
\end{itemize}

The findings emphasize the advantages of integrating domain-specific reasoning and advanced conflict resolution strategies in SMT solvers. For complex constraint satisfaction problems, such as large-scale Sudoku, SMT solvers offer unparalleled scalability and efficiency. These attributes make them the preferred choice for 
solving such challenges. 

The results underscore the need for continued development and refinement of SMT techniques to further enhance their applicability and performance across diverse problem domains.

\section{Conclusion}
This paper evaluated the performance of modern SMT solvers, such as Z3 and CVC5, against an older SMT solver in DPLL(T) and a classical SAT solver in DPLL on 25x25 Sudoku puzzles. The results demonstrate that modern SMT solvers significantly outperform SAT solvers on constraint satisfaction problems. Notably, Z3 and CVC5 achieved perfect success rates, significantly lower solving times, and markedly greater propagation efficiency. These findings exemplify the efficiency of modern SMT solvers, and highlight their utility on large-scale constraint satisfaction problems.

\section{Future Work}

In this paper, we evaluated the performance of various solvers using standard configurations to solve 25x25 Sudoku puzzles. While the results demonstrate the clear advantages of modern SMT solvers like CVC5 and Z3, there are several promising avenues for future work to enhance solver performance and broaden their applicability.

\subsection{Fine-Tuning SMT Solvers}

Future work could focus on fine-tuning SMT solvers like CVC5 and Z3 for specific problem domains. While the current study used fairly standard configurations, SMT solvers offer a range of parameters that can be optimized to achieve better performance on particular types of problems. For example:
\begin{itemize}
    \item Customizing heuristics for branching and variable selection to improve solving efficiency.
    \item Tailoring conflict resolution strategies to better handle specific classes of constraints.
    \item Exploring advanced preprocessing techniques to simplify problem instances before solving.
\end{itemize}
Such targeted optimizations could enable SMT solvers to outperform their standard implementations for specific problem sets, making them even more competitive in diverse application areas.

\subsection{Development of Hybrid Solvers}

Hybrid solvers that integrate the strengths of both SAT and SMT solvers present another exciting direction for future research. Combining the speed and simplicity of SAT solvers for handling Boolean constraints with the advanced theory-reasoning capabilities of SMT solvers could yield significant performance improvements. Potential areas of exploration include:
\begin{itemize}
    \item Designing hybrid frameworks that dynamically switch between SAT and SMT solving strategies based on the characteristics of the problem.
    \item Leveraging SAT solvers for initial preprocessing or rapid conflict detection, followed by SMT solvers for detailed theory reasoning.
    \item Exploring modular solver architectures that allow seamless integration of specialized solvers for different constraint types.
\end{itemize}
Such approaches could lead to solvers that are both efficient and versatile, capable of tackling a wider range of complex problems.

\subsection{Improving Modern SMT Solvers}

There is substantial scope for improving the core components of modern SMT solvers. Key areas of focus include:
\begin{itemize}
    \item \textbf{Theory Propagation:} Enhancing the efficiency of theory-specific propagation to reduce redundant computations and accelerate solving.
    \item \textbf{Conflict Resolution Strategies:} Developing more sophisticated methods for analyzing conflicts and generating conflict clauses to guide future search steps effectively.
    \item \textbf{Heuristics for Branching and Variable Selection:} Refining heuristics to make smarter decisions during the search process, thereby reducing the search space and solving time.
\end{itemize}

\subsection{Machine-Assisted Theorem Proving}

One of the most promising directions for future work lies in the integration of machine learning (ML) techniques and large language models (LLMs) into solvers, creating a new paradigm of machine-assisted theorem proving. By combining the high-level reasoning capabilities of ML with the low-level precision and efficiency of SAT/SMT solvers, this approach could unlock novel solutions to complex problems. Potential advancements include:
\begin{itemize}
    \item \textbf{Learning-Based Heuristics:} Using machine learning models to dynamically learn and predict effective branching and variable selection strategies based on historical solving data.
    \item \textbf{Online Learning for Sets of Related Problems:} Developing methods that allow solvers to continuously learn and adapt while solving a series of related problems can improve solver performance on the fly, enabling more efficient more efficient solving as new problems are encountered.
    \item \textbf{Integrating LLMs:} Combining the high level abstract reasoning capabilities of LLMs with the low level reasoning capabilities of SMT solvers is a fairly new approach that has shown the potential to unlock more efficent solutions to complex reasoning tasks.
\end{itemize}

This integration could lead to solvers that not only perform better on traditional benchmarks but also extend their utility to emerging fields like automated theorem proving, formal methods, and AI-driven scientific discovery.

\subsection{Exploring New Application Domains}

Finally, further research can focus on applying these solvers to novel domains beyond Sudoku, such as hardware and software verification, formal verification of AI systems, and even proving mathematical theorems through formal methods. Understanding how solver techniques generalize to diverse applications will provide valuable insights into their versatility and potential for broader impact.

\section{Acknowledgements}

This project was completed as the final project for COSC-241: Artificial Intelligence at Amherst College during the Fall 2024 semester, taught by Professor Andrew Wu. We extend our gratitude to Professor Wu for his 
support, insightful lectures, and guidance throughout the course, which provided the foundational knowledge and inspiration for this work.

We are also grateful to the broader academic and open-source communities for the development and maintenance of tools like PySAT, Z3, and CVC5, which were central to the implementation and analysis in this project. Their contributions enable students and researchers to explore and innovate in the fields of automated reasoning and artificial intelligence.

\bibliographystyle{IEEEtran}
\bibliography{references}

\begin{thebibliography}{1}
\providecommand{\url}[1]{#1}
\csname url@samestyle\endcsname
\providecommand{\newblock}{\relax}
\providecommand{\bibinfo}[2]{#2}
\providecommand{\BIBentrySTDinterwordspacing}{\spaceskip=0pt\relax}
\providecommand{\BIBentryALTinterwordstretchfactor}{4}
\providecommand{\BIBentryALTinterwordspacing}{\spaceskip=\fontdimen2\font plus
\BIBentryALTinterwordstretchfactor\fontdimen3\font minus \fontdimen4\font\relax}
\providecommand{\BIBforeignlanguage}[2]{{%
\expandafter\ifx\csname l@#1\endcsname\relax
\typeout{** WARNING: IEEEtran.bst: No hyphenation pattern has been}%
\typeout{** loaded for the language `#1'. Using the pattern for}%
\typeout{** the default language instead.}%
\else
\language=\csname l@#1\endcsname
\fi
#2}}
\providecommand{\BIBdecl}{\relax}
\BIBdecl

\bibitem{Nieuwenhuis2006}
\BIBentryALTinterwordspacing
R.~Nieuwenhuis, A.~Oliveras, and C.~Tinelli, ``Solving sat and sat modulo theories: From an abstract davis–putnam–logemann–loveland procedure to dpll(t),'' \emph{Journal of the ACM}, vol.~53, no.~6, pp. 937--977, 2006. [Online]. Available: \url{https://doi.org/10.1145/1217856.1217859}
\BIBentrySTDinterwordspacing

\bibitem{deMoura2008}
\BIBentryALTinterwordspacing
L.~M. de~Moura and N.~Bj{\o}rner, ``Z3: An efficient smt solver,'' in \emph{Tools and Algorithms for the Construction and Analysis of Systems (TACAS)}.\hskip 1em plus 0.5em minus 0.4em\relax Springer, 2008, pp. 337--340. [Online]. Available: \url{https://doi.org/10.1007/978-3-540-78800-3\_24}
\BIBentrySTDinterwordspacing

\bibitem{DBLP:conf/tacas/BarbosaBBKLMMMN22}
\BIBentryALTinterwordspacing
H.~Barbosa, C.~W. Barrett, M.~Brain, G.~Kremer, H.~Lachnitt, M.~Mann, A.~Mohamed, M.~Mohamed, A.~Niemetz, A.~N{\"{o}}tzli, A.~Ozdemir, M.~Preiner, A.~Reynolds, Y.~Sheng, C.~Tinelli, and Y.~Zohar, ``cvc5: {A} versatile and industrial-strength {SMT} solver,'' in \emph{Tools and Algorithms for the Construction and Analysis of Systems - 28th International Conference, {TACAS} 2022, Held as Part of the European Joint Conferences on Theory and Practice of Software, {ETAPS} 2022, Munich, Germany, April 2-7, 2022, Proceedings, Part {I}}, ser. Lecture Notes in Computer Science, D.~Fisman and G.~Rosu, Eds., vol. 13243.\hskip 1em plus 0.5em minus 0.4em\relax Springer, 2022, pp. 415--442. [Online]. Available: \url{https://doi.org/10.1007/978-3-030-99524-9\_24}
\BIBentrySTDinterwordspacing

\bibitem{2009Sudoku}
X.~Yuan-Hai, J.~Biao-Bin, L.~I. Yong-Zhuoadvisor, and S.~Hua-Fei, ``Sudoku puzzles generating: From easy to evil,'' \emph{Mathematics in Practice and Theory}, vol.~39, no.~21, pp. 1--7, 2009.

\end{thebibliography}

\end{document}